\documentclass[lettersize,journal]{IEEEtran}
\IEEEoverridecommandlockouts

\usepackage{amsmath,amssymb,amsfonts}
\usepackage{graphicx}
\usepackage{textcomp}
\usepackage{xcolor}
\usepackage{bm}
\usepackage{pifont}
\usepackage{amsthm}
\usepackage{mathrsfs}
\usepackage{enumerate}
\usepackage{multirow}
\usepackage{color}
\usepackage{threeparttable}
\usepackage{subfigure}
\usepackage{booktabs}
\usepackage{colortbl}
\usepackage{xcolor}
\definecolor{mygray}{gray}{0.8}

\usepackage[square, comma, sort&compress, numbers]{natbib}
\usepackage[pagebackref=false,breaklinks=true,colorlinks,bookmarks=false]{hyperref}
\usepackage{times}

\usepackage{array}
\usepackage{verbatim}
\usepackage{algorithm}
\usepackage{bbding}
\usepackage{algpseudocode}
\usepackage{lettrine}

\def\BibTeX{{\rm B\kern-.05em{\sc i\kern-.025em b}\kern-.08em
    T\kern-.1667em\lower.7ex\hbox{E}\kern-.125emX}}
\begin{document}

\title{MSSFC-Net: Enhancing Building Interpretation with Multi-Scale Spatial-Spectral Feature Collaboration\\

}
\author{Dehua Huo, Weida Zhan, Jinxin Guo, Depeng Zhu, Yu Chen, YiChun Jiang, Yueyi Han, Deng Han, and Jin Li
\thanks{This work was supported by the Development Program of the Department of Science and Technology of Jilin Province (Project Name: Development of Vehicle-Mounted Multi-Band Intelligent Fusion Enhanced Imaging and Processing System; Project No. 20240302029GX).

(Corresponding author: Weida Zhan) Dehua Huo, Weida Zhan, Jinxin Guo, Depeng Zhu, Yu Chen, Yichun Jiang and Yueyi Han are with the National Demonstration Center for Experimental Electrical, Changchun University of Science and Technology, Changchun 130022, China(e-mail: hdh@mails.cust.edu.cn; zhanweida@cust.edu.cn; guojinxin@mails.cust.edu.cn; dp.zhu@cust.edu.cn; chenyu@mails.cust.edu.cn; jiangyichun@mails.cust.edu.cn; hyy@mails.cust.edu.cn).}
\thanks{Deng Han is with the Jilin Province Zhixing IoT Research Institute Co.,Ltd., Changchun 130117, China(e-mail: handeng@live.com).}
\thanks{Jin Li is with the Beihang University, School of Instrumentation and Optoelectronic Engineering, Beihang University, Beijing 100191, China(e-mail: jl11269@buaa.edu.cn).}
}

\maketitle

\begin{abstract}
Building interpretation from remote sensing imagery primarily involves two fundamental tasks: building extraction and change detection. However, most existing methods address these tasks independently, overlooking their inherent correlation and failing to exploit shared feature representations for mutual enhancement. Furthermore, the diverse spectral, spatial, and scale characteristics of buildings pose additional challenges in jointly modeling spatial-spectral multi-scale features and effectively balancing precision and recall. The limited synergy between spatial and spectral representations often results in reduced detection accuracy and incomplete change localization.
To address these challenges, we propose a Multi-Scale Spatial-Spectral Feature Cooperative Dual-Task Network (MSSFC-Net) for joint building extraction and change detection in remote sensing images. The framework integrates both tasks within a unified architecture, leveraging their complementary nature to simultaneously extract building and change features. Specifically, a Dual-branch Multi-scale Feature Extraction module (DMFE) with Spatial-Spectral Feature Collaboration (SSFC) is designed to enhance multi-scale representation learning, effectively capturing shallow texture details and deep semantic information, thus improving building extraction performance.
For temporal feature aggregation, we introduce a Multi-scale Differential Fusion Module (MDFM) that explicitly models the interaction between differential and dual-temporal features. This module refines the network’s capability to detect large-area changes and subtle structural variations in buildings. Extensive experiments conducted on three benchmark datasets demonstrate that MSSFC-Net achieves superior performance in both building extraction and change detection tasks, effectively improving detection accuracy while maintaining completeness.

\end{abstract}


\section{Introduction}

\begin{figure}[t]
    \centering
    \includegraphics[width=1\linewidth]{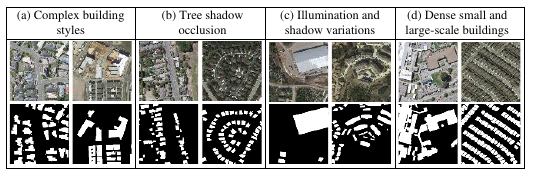}
\caption{Characteristics and challenges of building interpretation in remote sensing images, along with corresponding ground-truth segmentation mask example image pairs. Non-building areas represent the "background," and building areas represent the "foreground."}
\label{fig:1}
\end{figure}

\IEEEPARstart{B}{uilding} extraction and change detection are fundamental tasks in urban remote sensing, supporting applications such as disaster monitoring, urban planning, and land use management~\cite{dai2020object,mohammadian2023siamixformer,lu2024bi,gong2023context,lv2023spatial}. Although these tasks differ in objectives and processing pipelines, they share the same core target—buildings~\cite{mohammadian2023siamixformer,wang2024rsbuilding}. Both tasks require precise feature abstraction, laying the foundation for a joint modeling framework that could enhance overall performance through shared knowledge transfer.

Currently, most studies tackle building extraction and change detection separately. Building extraction methods focus on identifying building footprints from single-temporal images using advanced dual-branch networks, attention mechanisms, and cross-layer interactions~\cite{wang2024sdsnet,shen2023pbsl,sun2024g2ldie}. Transformer-based approaches further enhance global context modeling, enabling large-scale and high-precision building recognition~\cite{wang2022building,huang2023easy}.
Change detection (CD) methods aim to capture structural changes in buildings by comparing multi-temporal images, focusing on variations such as construction, demolition, and modifications~\cite{li2024ida}. Recent approaches leverage Siamese architectures, multi-scale features, and attention modules~\cite{pang2023sfgt,shen2024boosting,yang2024sedanet}. However, standard Transformer-based CD methods often struggle to model the complex spatial-spectral relationships critical for detecting subtle changes~\cite{bandara2022transformer,chen2021remote}.

Given the inherent correlation between the two tasks, joint learning offers the potential to improve both accuracy and completeness. Specifically, building extraction provides a baseline for change detection, while change detection supplies additional context for refining extraction results~\cite{wang2024sdsnet,wang2022building}. As illustrated in Fig.~\ref{fig:1}, fusing static and dynamic building information leads to more comprehensive and reliable outcomes.
Despite progress, challenges remain. Scale variations, complex backgrounds, and the spectral richness of remote sensing images complicate multi-task learning.   While attention mechanisms~\cite{shangguan2023attention,lv2023hierarchical} enhance feature extraction, existing methods often neglect explicit spatial-spectral modeling and rely on basic operations like feature concatenation or subtraction~\cite{zhang2023self,noman2024changebind,wang2023align}. This limits the ability to generate three-dimensional attention weights necessary for capturing the interplay between spatial and spectral features.

To address these limitations, we propose \textbf{MSSFC-Net}, a spatial-spectral collaborative dual-task network for building extraction and change detection. MSSFC-Net unifies both tasks within a Transformer-based framework, jointly modeling feature extraction and temporal differences. A dual-branch multi-scale feature extraction module (DMFE) with spatial-spectral feature collaboration (SSFC) is introduced to generate 3D spatial-spectral weights via Gaussian modeling, enabling precise multi-scale feature representation. Additionally, a multi-scale differential fusion module (MDFM) is designed to enhance temporal feature interaction and reduce noise in differential features.

The main contributions of this paper are summarized as follows:
\begin{itemize}
    \item We propose MSSFC-Net, a unified dual-task framework that jointly models building extraction and change detection, effectively leveraging task synergy for improved performance.
    \item A novel DMFE-SSFC module is designed to generate spatial-spectral 3-D attention weights without introducing extra parameters, achieving lightweight and efficient multi-scale feature extraction.
    \item We develop MDFM to refine dual-temporal feature fusion, enhancing the model's capability to capture both large-scale changes and fine-grained building variations while suppressing noise.
\end{itemize}

\begin{figure*}[t]
    \centering
    \includegraphics[width=1\linewidth]{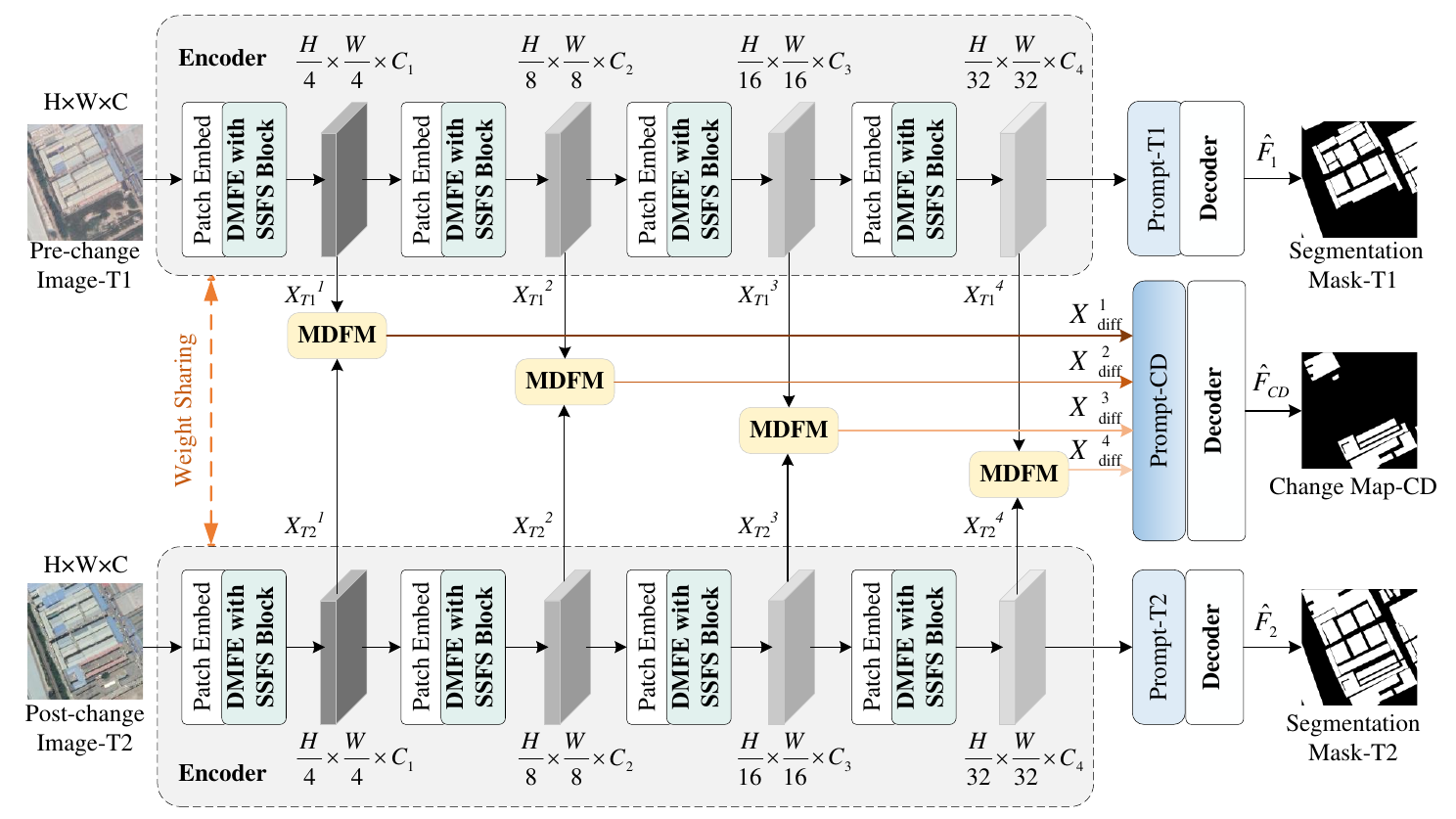}
\caption{The overall architecture of MSSFC-Net, a model that simultaneously processes dual-temporal imagery, is designed to extract both individual buildings and building changes.The model mainly consists of four components: a multi-scale contextual feature extraction module with spatial-spectral feature coordination, a multi-scale difference fusion module, a decoder that queries the corresponding semantic mask features based on task cues, and a segmentation head that generates the final segmentation results.}
\label{fig:2}
\end{figure*}

\section{Related Work}\label{sec:rw}

\subsection{Building Extraction in Remote Sensing}

Building extraction~\cite{hu2023robust} is a fundamental task in remote sensing, supporting urban planning and land use analysis~\cite{huang2023deep, wang2024rsbuilding}. Early CNN-based methods, such as FCN~\cite{yan2024enhancing} and U-Net~\cite{ronneberger2015u}, laid the foundation by modeling pixel-level segmentation. Extensions like UNet++~\cite{zhou2018unet++} and Attention U-Net~\cite{oktay2018attention} improved multi-scale feature fusion and object boundary precision.
Recent advances introduce transformer-based models for global context modeling. ViT~\cite{dosovitskiy2020image} and UFormer~\cite{wang2022uformer} enhance large-scale feature perception, while BuildFormer~\cite{wang2022building} further optimizes building extraction with hierarchical fusion. However, most existing methods overlook temporal dynamics, limiting their robustness in changing scenes. To address this, we propose a joint framework integrating building extraction and change detection.

\subsection{Change Detection and Temporal Feature Fusion}

Change detection identifies scene changes from multi-temporal imagery. Siamese networks~\cite{daudt2018fully} and their variants~\cite{fang2021snunet, ying2024dgma} enable paired feature learning but face challenges in capturing complex temporal dependencies.
Transformer-based approaches~\cite{chen2021remote, li2022transunetcd, gedara2022transformer} leverage global attention for robust temporal modeling. Recent methods such as APD~\cite{wang2023align} and IDA-SiamNet~\cite{li2024ida} refine difference representation through alignment and adaptive fusion. Despite progress, most models rely on static fusion or simple difference operations, limiting their ability to capture fine-grained changes.
Motivated by recent advances in progressive diffusion~\cite{shen2023advancing} and conditional generation~\cite{shen2024imagdressing, shen2024imagpose}, we introduce an adaptive temporal interaction module to enhance bitemporal feature fusion and change localization.

\subsection{Attention Mechanisms and Conditional Modeling}

Attention mechanisms have become essential in remote sensing tasks for capturing long-range dependencies~\cite{vaswani2017attention}. Channel and spatial attention models~\cite{hu2018squeeze, woo2018cbam} improve feature discrimination, while spectral-spatial attention~\cite{lei2021difference, li2023cbanet} enhances change detection.
Recent trends explore advanced conditioning and memory mechanisms. External attention~\cite{guo2022beyond} and bipartite-aware learning~\cite{shen2023pbsl} improve feature matching, while rich-contextual conditional diffusion~\cite{shen2024boosting} demonstrates the potential of integrating dynamic scene priors.
Building on these insights, our method incorporates adaptive attention and temporal conditioning to improve multi-task performance, enabling robust building extraction and fine-grained change detection in complex urban environments.

\section{Proposed Method}\label{sec:method} 
\subsection{Overview}

The overall architecture of MSSFC-Net is illustrated in Fig.~\ref{fig:2}. Our framework adopts a weight-sharing Siamese structure comprising three key components: an encoder, a decoder, and a multi-task segmentation head. The segmentation head simultaneously produces three binary masks, corresponding to building segmentation at two time points and building change detection.
Given a pair of pre-aligned bi-temporal images (T1 and T2), MSSFC-Net first extracts rich spatial-spectral features through a dual-branch encoder. The encoder integrates a MSFF and a SSFC. Specifically, the MSFF captures hierarchical multi-scale representations, while the SSFC enhances spatial-spectral dependencies to generate more robust feature embeddings.
To effectively model temporal differences, the extracted features are further processed by the MDFM, which explicitly aggregates dual-temporal features and emphasizes change-relevant information, thereby improving change detection accuracy.
In the decoder stage, an efficient multi-task decoder is designed to unify building extraction and change detection. We introduce three learnable query embeddings, each corresponding to a specific task branch. These task-specific queries interact with the bi-temporal feature maps through a cross-attention mechanism, selectively attending to task-relevant information. The resulting features are then introduced into the segmentation head to generate precise building masks and change maps.
Overall, MSSFC-Net achieves joint modeling of building extraction and change detection by leveraging spatial-spectral collaboration and temporal difference fusion, enabling robust performance across both tasks.

\begin{figure*}[t]
    \centering
    \includegraphics[width=1\linewidth]{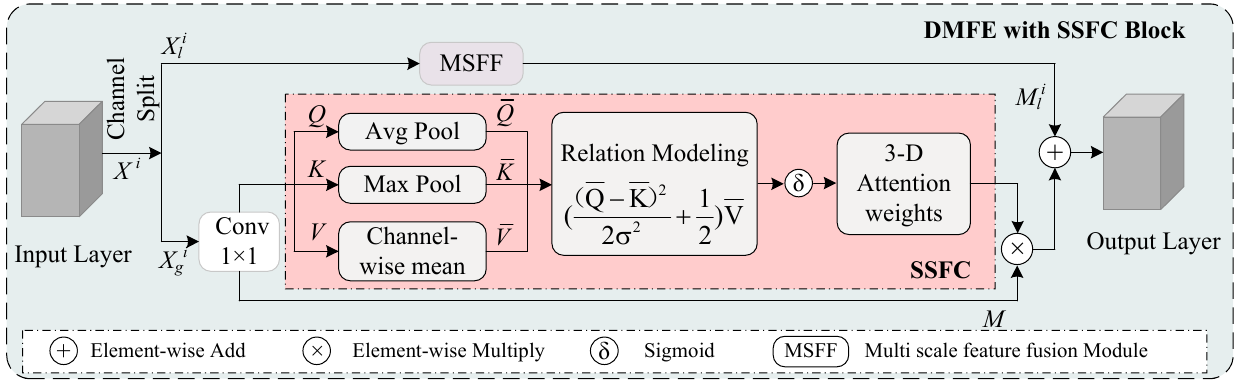}
\caption{The DMFE with SSFC divides the input features by channels and introduces them into two separate parallel branch context aggregators to obtain multi-scale information and spatial-spectral feature information.The SSFC strategy generates 3-D attention weights on the feature maps through heuristic computation, without the need for any learnable parameters. Additionally, to improve the efficiency of 3-D attention, relational modeling is performed on the $\bar{Q}$ , $\bar{K}$ and  $\bar{V}$ tokens within a channel subset (C/4), enhancing the edges and internal details of dynamic targets in remote sensing images.}
\label{fig:3}
\end{figure*}

\begin{figure}[t]
    \centering
    \includegraphics[width=1\linewidth]{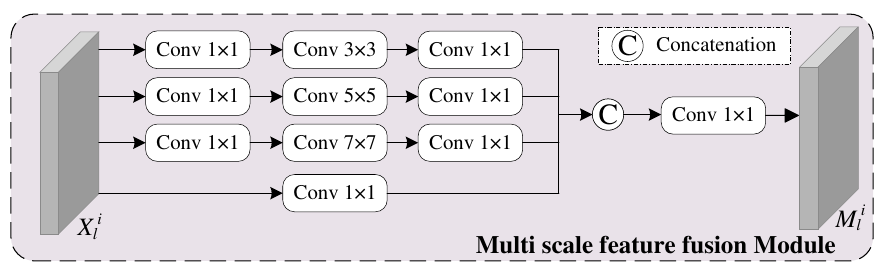}
\caption{MSFF structure. The multi-scale features are extracted by combining multi-branch structures and convolution kernels of different scales. The feature information from different branches is aggregated through channel fusion, enabling the comprehensive extraction of features across multiple scales.}
\label{fig:4}
\end{figure}

\subsection{Dual-Branch Multi-Scale Feature Extractor with Spatial-Spectral Feature Collaboration}

Traditional multi-scale feature extraction methods often introduce redundant computational overhead due to repeated convolutions or heavy self-attention mechanisms. To address this, we propose a Dual-Branch Multi-Scale Feature Extraction module enhanced with Spatial-Spectral Feature Collaboration (SSFC), which efficiently captures multi-scale contextual information while maintaining accuracy. Inspired by the parallel design of Inception-v1~\cite{szegedy2015going}, we divide the input channels into two groups, each processed by distinct contextual aggregation branches, as illustrated in Fig.~\ref{fig:3}.

Given input feature $X^i \in \mathbb{R}^{H^i \times W^i \times C^i}$ at stage $i$, we split it channel-wise into $X^i_{g} \in \mathbb{R}^{H^i \times W^i \times \frac{C^i}{2}}$ for the two branches.

\textbf{Multi-Scale Contextual Branch:} 
This branch captures contextual information at multiple scales via parallel convolutions with varying kernel sizes,as illustrated in Fig.~\ref{fig:4}. The operations are defined as:
\begin{equation}
\begin{aligned}
    M^i_l = \text{Conv}_{1\times1}\Big(&\text{Cat}\big(\text{Conv}_{1\times1}(X^i_l), \\
    &\text{Conv}_{1\times1}(\text{Conv}_{3\times3}(\text{Conv}_{1\times1}(X^i_l))), \\
    &\text{Conv}_{1\times1}(\text{Conv}_{5\times5}(\text{Conv}_{1\times1}(X^i_l))), \\
    &\text{Conv}_{1\times1}(\text{Conv}_{7\times7}(\text{Conv}_{1\times1}(X^i_l))) \big) \Big).
\end{aligned}
\end{equation}
where $\text{Cat}(\cdot)$ denotes channel-wise concatenation. This design captures rich multi-scale semantics without significantly increasing parameters.

\textbf{Spatial-Spectral Feature Cooperation (SSFC) Branch:} 
Direct fusion of global and local features often overlooks spatial-spectral dependencies. To mitigate this, we design the SSFC branch that explicitly models such dependencies. Specifically, a $1\times1$ convolution generates features $M$, $Q$, $K$, and $V$, where $(Q, K, V)$ represent the query, key, and value for attention computation, and $M$ aggregates multi-channel information.

Average pooling (stride 2, $3\times3$ kernel) and max pooling (stride 2, $2\times2$ kernel) are applied to $Q$ and $K$ to produce $\bar{Q}$ and $\bar{K}$. The channel-wise average $\bar{V}$ is computed from $X^i_g$.

Following~\cite{lei2023ultralightweight}, we extend Nadaraya-Watson kernel regression to tensor operations and design a Gaussian kernel-based spatial-spectral fusion:
\begin{equation}
Y = \text{Sigmoid} \left( \frac{(\bar{Q} - \bar{K})^2}{2\sigma^2} + \frac{1}{2} \right) \times \bar{V}.
\end{equation}
where $\sigma^2$ denotes the variance of channel dimensions, reflecting the contextual richness. The Sigmoid function normalizes attention weights, promoting better feature refinement. Unlike conventional attention mechanisms~\cite{woo2018cbam,fu2019dual}, our SSFC achieves fine-grained enhancement of target boundaries and internal structures without introducing additional learnable parameters.

\subsection{Multi-Scale Differential Fusion Module (MDFM)}

\begin{figure}[t]
    \centering
    \includegraphics[width=1\linewidth]{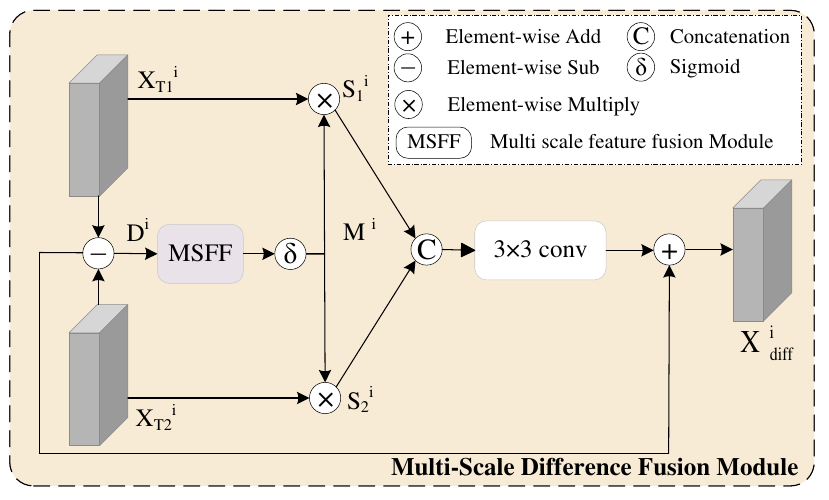}
\caption{The MDFM structure. It is used to fuse the features obtained from dual-temporal images, generating differential features with contextual information. After generating the initial differential features \( D^{i} \), a multi-scale feature learning mechanism is enhanced to fuse the dual-temporal features.}
\label{fig:5}
\end{figure}

To enhance temporal difference modeling, we design the Multi-Scale Differential Fusion Module (MDFM) that aggregates multi-level features while capturing contextual change cues, as illustrated in Fig.~\ref{fig:5}. Given dual-temporal features $X_{T_1}^i$ and $X_{T_2}^i$ from stage $i$, initial differential features $D^i$ are computed as:
\begin{equation}
D^i = \left| X_{T_1}^i - X_{T_2}^i \right|.
\end{equation}
where $|\cdot|$ denotes element-wise absolute difference.

Each $D^i$ undergoes multi-scale processing via convolutional filters with varying kernel sizes. Channel attention weights $M^i$ are computed as:
\begin{equation}
M^i = \sigma(M_l^i).
\end{equation}
where $\sigma(\cdot)$ is the Sigmoid activation.

Weighted dual-temporal features are obtained by:
\begin{equation}
\begin{aligned}
S_i^1 &= M^i \times X_{T_1}^i, \\
S_i^2 &= M^i \times X_{T_2}^i.
\end{aligned}
\end{equation}

The final fused differential feature $X_{\text{diff}}^i$ is derived via residual learning and $3\times3$ convolution:
\begin{equation}
X_{\text{diff}}^i = D^i + \text{Conv}_{3\times3} \left( \text{Stack}(S_i^1, S_i^2) \right).
\end{equation}
where $\text{Stack}(\cdot)$ denotes channel concatenation. This design ensures robust fusion of temporal differences, preserving subtle changes and mitigating noise interference.

\subsection{Segmentation Head}  
We utilize three query embeddings to encode task-specific semantic features. These three queries correspond to the building extraction tasks at two different time points and the associated change detection task, denoted as 
\( e_{1} \) , \( e_{2} \) , and \( e_{cd} \), respectively. The three query embeddings model spatiotemporal relationships for specific tasks through cross-attention mechanisms. The detailed procedure is as follows: The three prompts are concatenated along the sequence dimension, and the predefined position encoding mapping and time encoding mapping are concatenated to the corresponding task features to form the image feature \( E^{i} \) at the i-th layer. 
The dual-temporal image feature is obtained by flattening the two-dimensional features into a one-dimensional feature sequence:
\begin{equation}
\begin{aligned}
E^i &= \Phi_{\text{Cat}}(e_1, e_2, e_{\text{cd}}). \\
\end{aligned}
\end{equation}
The symbol $\Phi_{\text{Cat}}$  represents the operation of concatenating vectors along the sequence dimension.

Task-specific dense feature representations are obtained through the multi-head self-attention layer, multi-head cross-attention layer, and linear projection layer, which are used to interpret the corresponding masks:

\begin{equation}
\begin{aligned}
E^i &= \Phi_{\text{S-attn}}(E^{i-1}) ,\\ 
E^i &= \Phi_{\text{C-attn}}(E^i, T^i) ,\\
E^i &= \Phi_{\text{mlp-proj}}(E^i) .\\
\end{aligned}
\end{equation}
where $\Phi_{\text{S-attn}}$ represents the standard multi head self attention layer, $\Phi_{\text{C-attn}}$ represents a multi head cross attention layer, $\Phi_{\text{mlp-proj}}$ refers to a linear projection layer.

A linear projection layer is used to reduce the dimensionality of the task-specific embeddings in the final layer:

\begin{equation}
\begin{aligned}
\hat{E} &= \Phi_{\text{e-proj}}(E^4), \\
\hat{e}_1, & \hat{e}_2, \hat{e}_\text{cd} = \hat{E}.\\
\end{aligned}
\end{equation}
where $\Phi_{\text{e-proj}}$ refers to the linear projection layer, which is used to reduce the dimensionality of the task-specific embeddings at the final layer. 

Einstein summation is employed to obtain the explanatory results for various tasks:
\begin{equation}
\begin{aligned}
\hat{m}_1 &= \Phi_{\text{e-sum}}(\hat{F}_1, \hat{e}_1). \\
\end{aligned}
\end{equation}

The segmentation mask $\hat{m}_1$ is obtained by $\Phi_{\text{e-sum}}$, which computes a linear weighted sum of 
$\hat{F}_1$ and $\hat{e}_1$ . It is important to note that the derivation processes for 
$\hat{m}_2$ and $\hat{m}_\text{cd}$ follow the same procedure.
\section{Experiment and Analysis}\label{sec:exp}  

\subsection{Experimental Datasets and Settings}
In this paper, we conduct experiments on three different datasets: a remote sensing segmentation dataset, a change detection dataset, and a dual-label dataset. These datasets are WHU~\cite{ji2018fully}, LEVIR-CD~\cite{zhang2020deeply}, and BANDON~\cite{pang2023detecting}.

\textbf{\emph{WHU}}~\cite{ji2018fully}:The WHU Building Dataset comprises two subsets: one for satellite imagery and another for aerial imagery. This study utilizes the aerial imagery subset, which contains 8,189 images with a spatial resolution of 0.3 m/pixel. The dataset is partitioned into 4,736 training, 1,036 validation, and 2,416 test images. Covering an area exceeding 450 km², the aerial subset includes approximately 22,000 annotated buildings.

\textbf{\emph{LEVIR-CD}}~\cite{zhang2020deeply}:This dataset consists of 637 bi-temporal high-resolution remote sensing image pairs (0.5 m/pixel), each sized 1,024 × 1,024 pixels. Binary change annotations focus on building construction and demolition, with imagery sourced from Google Earth spanning a temporal range of 5 to 14 years. Following official guidelines, the dataset is divided into 445 pairs for training, 64 for validation, and 128 for testing.

\textbf{\emph{BANDON}}~\cite{pang2023detecting}:The BANDON dataset features high-resolution imagery (0.6 m/pixel) with dual annotations for semantic segmentation and building change detection. Data sources include Google Earth, Microsoft Virtual Earth, and ArcGIS, covering six representative Chinese cities: Beijing, Shanghai, Wuhan, Shenzhen, Hong Kong, and Jinan. The dataset contains 1,689 training pairs, 202 validation pairs, and 392 test pairs, with original image dimensions of 2,048 × 2,048 pixels. For experimental purposes, all images were cropped to 512 × 512 pixels.

\subsection{Evaluation Protocol and Metrics} 
To evaluate the performance of the proposed method, we conducted a comprehensive assessment of the experimental results using three widely used evaluation metrics: Intersection over Union (IoU), Precision (P), Recall (R), and F1-score (F1). These metrics are defined as follows:
\begin{equation}
\begin{aligned}
\text{IoU} &= \frac{\text{TP}}{\text{TP}+\text{FP}+\text{FN}} , \\
\text{P} &= \frac{\text{TP}}{\text{TP}+\text{FP}} , \\
\text{R} &= \frac{\text{TP}}{\text{TP}+\text{FN}} , \\
\text{F1} &= \frac{2 \times \text{P} \times \text{R}}{\text{P}+\text{R}} .\\
\end{aligned}
\end{equation}
where TP, FP, and FN denote true positives, false positives, and false negatives, respectively~\cite{li2022transunetcd}.
\subsection{Implementation Details} 

To accommodate heterogeneous tasks within a unified framework and meet the requirements of both single-image building extraction and dual-image change detection tasks, we adopted a dual-image input scheme as a compromise.
In this study, the experiments were conducted using the PyTorch framework and an NVIDIA GeForce RTX 4090 GPU for training. The input image size was fixed at 512 × 512. Random flipping and cropping techniques were employed to augment the images of the three datasets.
The Adam optimization algorithm was used to optimize the model, with the momentum set to 0.99 and the weight decay set to 0.0005. During the training process, the batch size was set to 16, and the learning rate was set to 0.0001.
For the loss function, we utilized pixel-wise cross-entropy loss to measure the performance of the network during training.

\subsection{Comparison with State-of-the-art Methods} 

We evaluated the effectiveness of the MSSFC-Net method by comparing it with several state-of-the-art building extraction and change detection methods. The selected comparison methods encompass both semantic segmentation and building extraction techniques, including: U-Net~\cite{ronneberger2015u}, DeepLabv3+~\cite{chen2018encoder}, HRNet~\cite{wang2020deep}, SegFormer~\cite{xie2021segformer}, MAP-Net~\cite{zhu2020map}, BOMSC-Net~\cite{zhou2022bomsc}, BuildFormer~\cite{wang2022building}, BCTNet~\cite{xu2023bctnet}, DSAT-Net~\cite{zhang2023dsat}, SDSNet~\cite{wang2023sdsnet}, SAU-Net~\cite{chen2024sau} and CSA-Net~\cite{yang2024csa}. For change detection, the methods include: FC-Siam-conc~\cite{daudt2018fully}, FC-Siam-diff~\cite{daudt2018fully}, BIT~\cite{chen2021remote}, ChangeFormer~\cite{bandara2022transformer}, TransUNetCD~\cite{li2022transunetcd}, ChangerEx~\cite{fang2023changer}, SGNet~\cite{feng2024sgnet}, DGMA$^2$-Net~\cite{ying2024dgma}, LCD-Net~\cite{liu2025lcd}, and IDA-SiamNet~\cite{li2024ida}. The performance metrics of these methods are based on the officially published results and those obtained from our reimplementation using PyTorch.
\subsubsection{Quantitative and Qualitative Comparison of Building Extraction}

\begin{table}[t]
\centering
\caption{Comparison of Building Extraction Performance of Different Methods on the WHU Test Set. Our MSSFC-Net effectively leverages and integrates shallow texture detail information and deep semantic localization information, achieving outstanding performance in terms of Precision (P), Intersection over Union (IoU), and F1 score on the WHU(\%) dataset. The best results are highlighted in bold.}
\label{tab:whu}
\begin{minipage}{0.48\textwidth} 
\resizebox{\textwidth}{!}{
    \begin{tabular}{c|c|cccc}
    \toprule
    \multirow{1}{*}{Method} & \multirow{1}{*}{Year} & \multicolumn{1}{c}{P} & \multicolumn{1}{c}{R} & \multicolumn{1}{c}{IoU} & \multicolumn{1}{c}{F1} \\ 
    \midrule   
    U-Net~\cite{ronneberger2015u} & 2015 
    &93.88 &94.92 &89.15 &94.40  
    \\
    DeepLabv3+~\cite{chen2018encoder} & 2017
    &94.05 &93.78 &89.03 &93.91
    \\
    HRNet~\cite{wang2020deep} & 2020 
    &94.89 &95.34 &89.87 &95.11 
    \\
    SegFormer~\cite{xie2021segformer} & 2021 
    &94.27 &94.83 &89.32 &94.55
    \\ 
    MAP-Net~\cite{zhu2020map} & 2021 
    &94.36 &95.39 &90.35 &94.87
    \\ 
    BOMSC-Net~\cite{zhou2022bomsc} & 2022 
    &95.14 &94.50 &90.15 &94.80  	
    \\
    BuildFormer~\cite{wang2022building}& 2022 
    &95.86 &95.21 &91.20 &95.53  
    \\ 
    BCTNet~\cite{xu2023bctnet} & 2023 
    &95.47 &95.27 &91.15 &95.37   
    \\ 
    DSAT-Net~\cite{zhang2023dsat} & 2023 
    &96.02 &94.95 &91.74 &95.48    	 
    \\
    SDSNet~\cite{wang2023sdsnet}  & 2024 
    &95.29 &94.38 &90.17 &94.83  
    \\
    SAU-Net~\cite{chen2024sau} & 2024  
    &95.61 &95.11 &91.12 &95.37
    \\
    CSA-Net~\cite{yang2024csa} &2024 
    &95.82 &\textbf{95.51} &89.83 &95.66 
    \\   
    \midrule  
    MSSFC-Net (Ours) & -
    & \textbf{96.30} & 95.34 & \textbf{92.27} & \textbf{95.82}
    \\    
    \bottomrule
    \end{tabular}
}
\end{minipage}
\end{table}

In this section, we present the performance of the MSSFC-Net method for building extraction tasks and compare quantitative metrics, as shown in Tabel  \ref{tab:whu}. Compared to U-Net, DeepLabv3+, HRNet, SegFormer, MAP-Net, BuildFormer, and DSAT-Net, our method achieves improvements in IoU of 3.12\%, 3.24\%, 2.4\%, 2.95\%, 1.92\%, 1.07\%, and 0.53\%, respectively. Unlike the sequential concatenation of attention and convolutional layers, our approach integrates both layers into a single module to simulate spatial-spectral dependencies. Moreover, our method exhibits notable advantages in capturing fine-grained details and discriminating building boundaries with enhanced precision.

\begin{figure*}[t]
    \centering
    \includegraphics[width=1\linewidth]{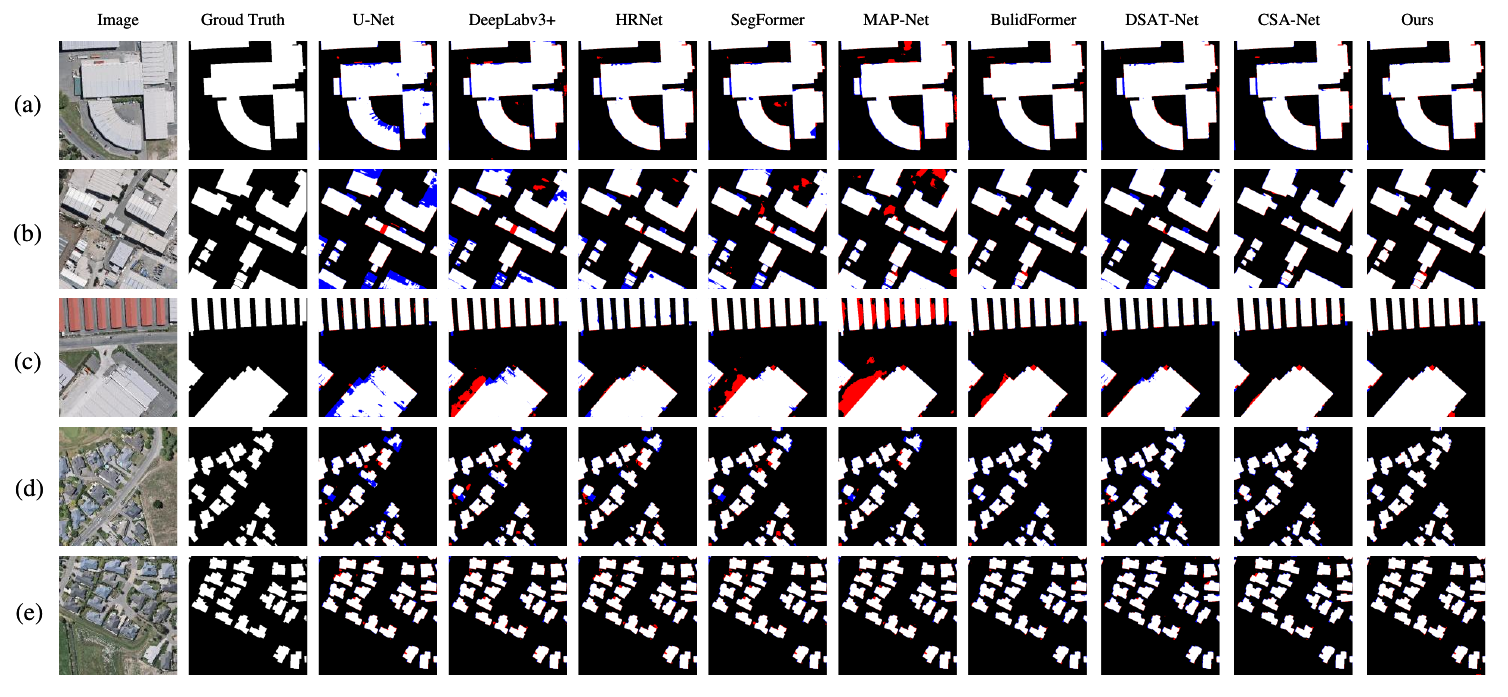}
\caption{Visualizes a comparison of the results on the WHU test set, where red represents false positives (FP), blue represents false negatives (FN), white denotes true positives, and black denotes true negatives.We present comparison with the best eight existing extraction methods in literature, whose codebases are publicly available. The results shows that our method is able to more accurately delineate building extraction areas.}
\label{fig:6}
\end{figure*}

To identify the causes of this result, we visualized the scenarios in the test set, as presented in Fig. \ref{fig:6}, which illustrates three types of buildings: dense buildings, large-scale buildings, and irregular-shaped buildings.Compared to models that solely employ multi-scale methods, MSSFC-Net incorporates the SSCF spectral-spatial collaboration strategy on top of multi-scale techniques, enabling it to capture more complex building features. Unlike the simple concatenation of low-level features seen in DeepLabv3+, the SSCF strategy filters cluttered backgrounds and enhances edges and internal details of changed targets in remote sensing images.

As illustrated in Fig .\ref{fig:6}(a)-(c), MSSFC-Net effectively recognizes architectural morphological features at different spatial scales through its multi-scale feature fusion mechanism. It accurately extracts geometric contours while preserving the main structure of the building. Experimental comparisons demonstrate that mainstream models suffer from significant misclassification in the boundary regions between buildings. In contrast, MSSFC-Net reduces the misidentification rate of non-building objects through its multi-level feature decoupling strategy. 
As shown in Fig .\ref{fig:6}(d)-(e) for irregular-shaped buildings, conventional algorithms exhibit blurred boundaries. In contrast, the DMFE module integrated into MSSFC-Net significantly enhances the feature discrimination capability for building targets through its spectral-spatial feature coupling analysis mechanism. Results indicate that in spectrally confusing scenarios, the DMFE module achieves a substantial improvement in building recognition accuracy. The core lies in the dynamic weight allocation mechanism of the SSFC fusion strategy: spatial attention-guided spectral feature reconfiguration effectively suppresses environmental interference features while boosting the feature response intensity of building components. Visual comparisons confirm that this module maintains building contour integrity while reducing missegmentation rates for complex roof structures. This improvement stems from the module's deep modeling of spatial contextual relationships for building targets to preserve precise recognition rates.

\begin{figure*}[t]
    \centering
    \includegraphics[width=1\linewidth]{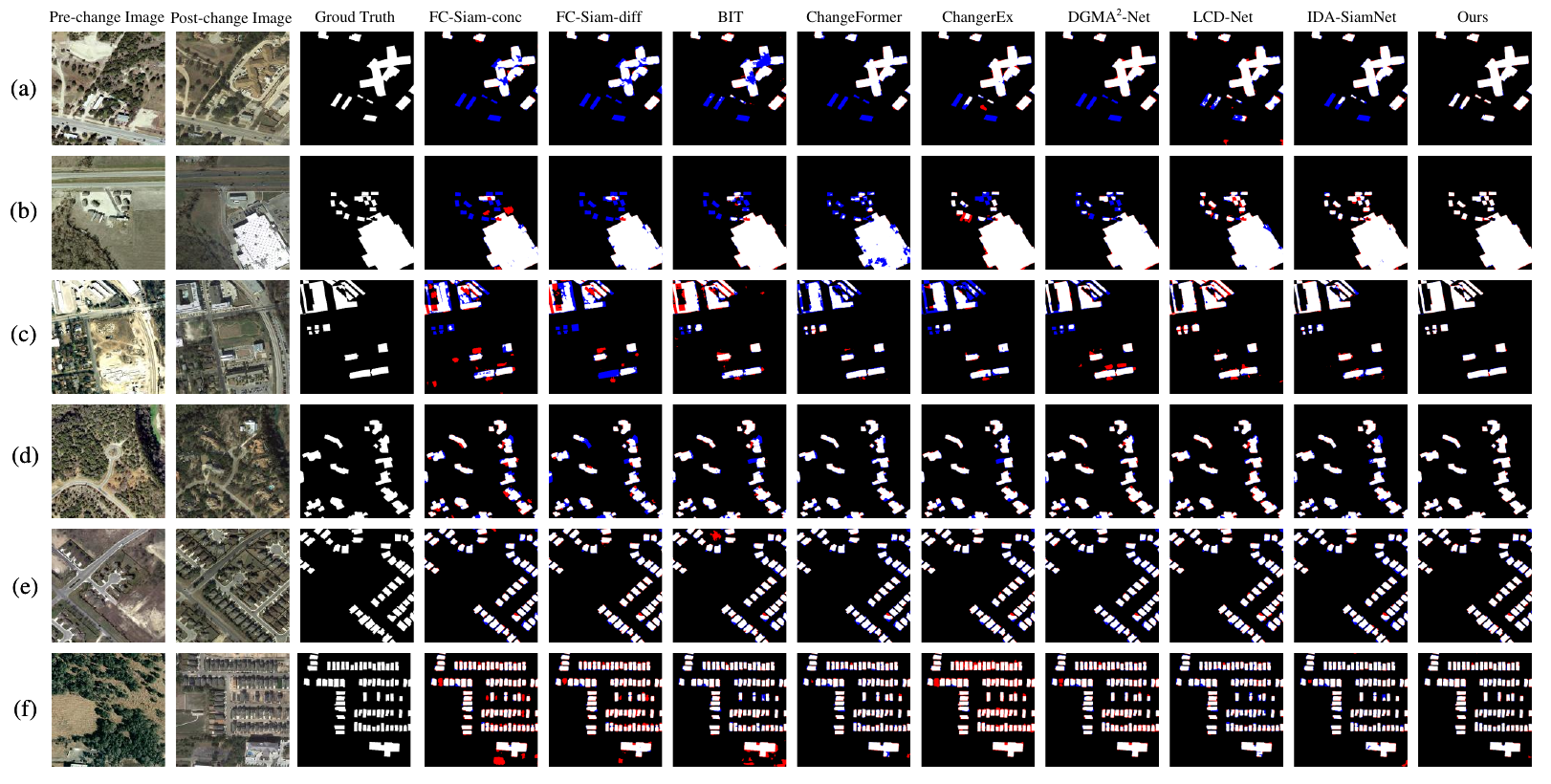}
\caption{Visualizes a comparison of the results on the LEVIR-CD test set, where red represents false positives (FP), blue represents false negatives (FN), white denotes true positives, and black denotes true negatives.We present comparison with the best eight existing change detection methods in literature, whose codebases are publicly available. The results shows that MSSFC-Net is more suitable for detecting change areas with clearly defined boundaries in pre- and post-change images.}
\label{fig:7}
\end{figure*}

\begin{figure*}[htbp]  
\centering
\includegraphics[width=\textwidth]{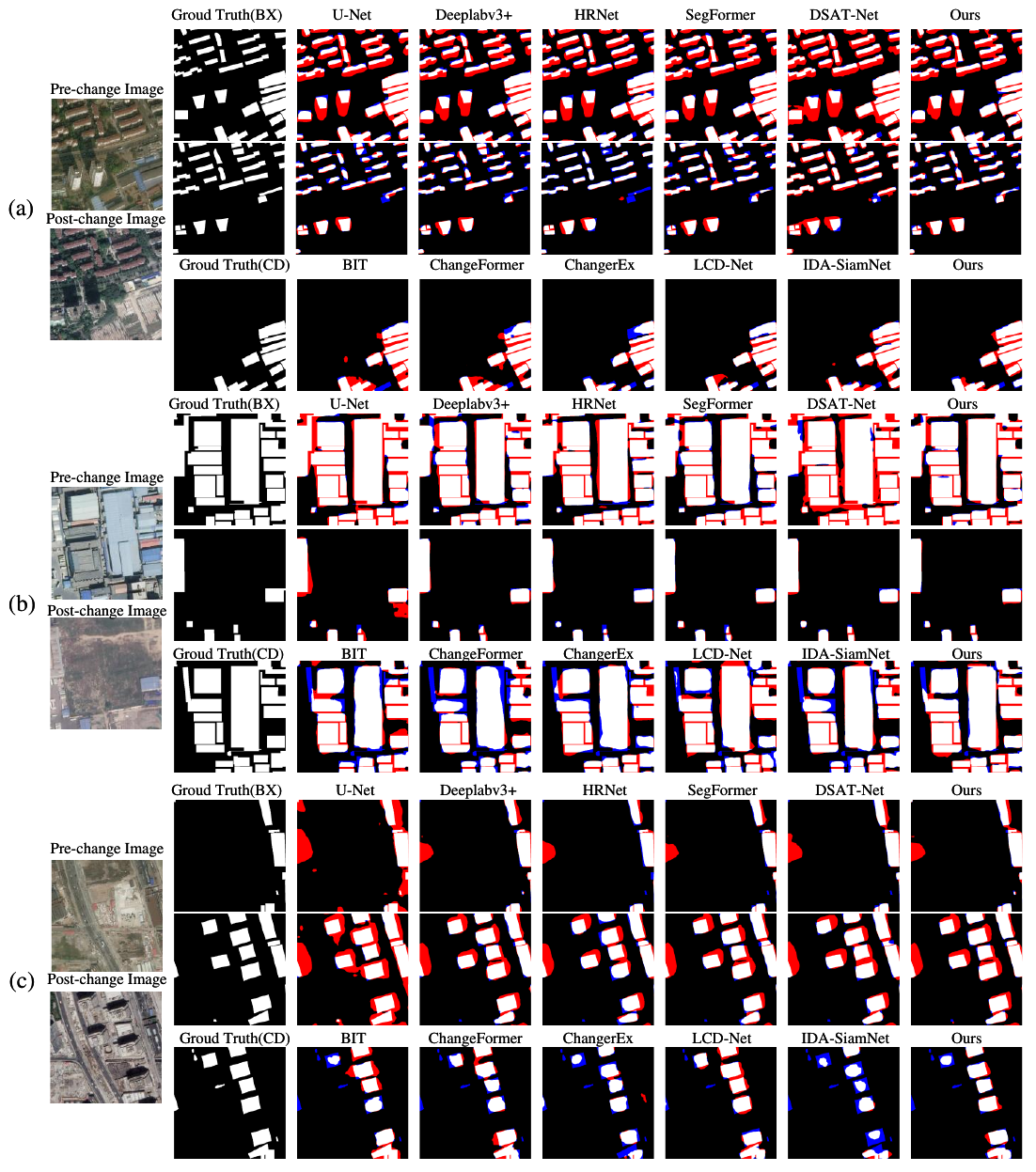}  
\caption{Visualizes a comparison of the results on the BANDON test set, where red represents false positives (FP), blue represents false negatives (FN), white denotes true positives, and black denotes true negatives.}
\label{fig:8}
\end{figure*}

\subsubsection{Quantitative and Qualitative Comparison of Change Detection}

\begin{table}[t]
\centering
\caption{ Performance comparison of different methods on the LEVIR-CD test set for change detection. Our MSSFC-Net effectively leverages the relationship between differential features and bi-temporal features to integrate the bi-temporal information, achieving outstanding performance in terms of Precision (P), Intersection over Union (IoU), and F1-score on the LEVIR-CD(\%) dataset. The best results are highlighted in bold.}
\label{tab:levircd}
\begin{minipage}{0.48\textwidth} 
\resizebox{\textwidth}{!}{
 \begin{tabular}{c|c|cccc}
    \toprule
    \multirow{1}{*}{Method} & \multirow{1}{*}{Year} & \multicolumn{1}{c}{P} & \multicolumn{1}{c}{R} & \multicolumn{1}{c}{IoU} & \multicolumn{1}{c}{F1} \\ 
    \midrule   
    FC-Siam-conc~\cite{daudt2018fully} & 2018 
    &88.33 &87.81 &78.68 &88.06    \\
    FC-Siam-diff~\cite{daudt2018fully} & 2018
    &91.32 &85.17 &78.79 &88.14 \\
    BIT~\cite{chen2021remote}& 2021 
    &89.30 &89.78 &81.06 &89.54  \\
    ChangeFormer~\cite{bandara2022transformer} & 2022
    &94.75 &86.31 &82.36 &90.33 \\ 
    TransUNetCD~\cite{li2022transunetcd} & 2022 
    &92.43 &89.82 &83.67 &91.11  \\ 
    ChangerEx~\cite{fang2023changer}& 2023 
    &91.47 &91.00 &83.88 &91.23  	  \\
    SGNet~\cite{feng2024sgnet} & 2024 
    &92.08 &89.92 &83.47 &90.99    \\ 
    DGMA$^2$-Net~\cite{ying2024dgma} & 2024 
    &93.56 &89.01 &83.87 &91.23      \\ 
    LCD-Net~\cite{liu2025lcd}  & 2024 
    &91.53 &\textbf{91.49} &84.34 &91.51    	    \\
    IDA-SiamNet~\cite{li2024ida}  & 2024 
    &93.21 &90.89 &85.28 &92.03     \\
    \midrule  
    MSSFC-Net (Ours) & -
    &\textbf{94.86} & 90.22 & \textbf{85.78} & \textbf{92.48}
  \\    
    \bottomrule
    \end{tabular}
}
\end{minipage}
\end{table}

In this section, we demonstrate the performance of the MSSFC-Net method in the construction of change detection tasks and compare quantitative metrics, as shown in Tabel  \ref{tab:levircd}. Compared to FC-Siam-conc, BIT, Change-Former, ChangerEx, DGMA$^2$-Net, LCD-Net, and IDA-SiamNet, our method achieves an improvement in IoU of 7.1\%, 4.72\%, 3.42\%, 1.9\%, 1.91\%, 1.44\%, and 0.5\%, respectively.This improvement stems from two key contributions: First, the dual-branch multi-scale feature extraction module generates hierarchical representations across spatial scales, thereby substantially enhancing the accuracy of building extraction. Second, through hierarchical feature interactions and effective exploitation of relationships between difference and temporal features, our method refines the capability to capture both large-scale regional transformations and subtle architectural modifications with unprecedented precision.

As shown in Fig .\ref{fig:7}(a)-(c), MSSFC-Net demonstrates significant accuracy advantages in building change detection tasks, with detection error significantly lower than that of comparative algorithms. In contrast, other network models exhibit substantial false positive regions. The visual comparison in Fig .\ref{fig:7}(d) clearly shows that MSSFC-Net can accurately capture the distribution characteristics of irregular buildings. In contrast, traditional methods exhibit obvious blurring effects in representing building boundaries and deformations. Especially, in the densely built-up samples shown in Fig .\ref{fig:7}(e)-(f), MSSFC-Net achieves complete extraction of dense building entities. Moreover, it demonstrates excellent performance in pixel-level edge detection. Experimental data further demonstrate that the building contours generated by the proposed network exhibit higher topological integrity and geometric fidelity. The completeness rate and accuracy rate of the recognition results show improvements compared to the baseline model, which fully validates the algorithm's precise recognition capability for multi-scale building targets in complex scenarios.

\subsubsection{Quantitative and qualitative comparison of the two tasks}

We conducted extensive evaluations using the BANDON dataset, which provides annotations for both building extraction and change detection tasks. The experimental results demonstrate that MSSFC-Net achieves the best performance in at least three metrics, with competitive IoU scores of 79.10\% and 56.38\% for each respective task, as shown in Tabel \ref{tab:BANDON}. The results highlight the complementary and representational capabilities between the two tasks.

\begin{table*}[t]
\renewcommand{\arraystretch}{0.9}  
\centering
\caption{Comparison of building extraction and change detection performance of different methods on the BANDON test set. This highlights the dual capability of our MSSFC-Net method: building extraction and change detection. It achieves outstanding performance across various metrics on the BANDON dataset. The best results are highlighted in bold.}
\label{tab:BANDON}
\resizebox{0.98\linewidth}{!}{
    \tiny
    \begin{tabular}{c|c|cccc|cccc}
    \toprule
    \multirow{2}{*}{Method}& \multirow{2}{*}{Year} & \multicolumn{4}{c|}{Building Extraction (\%)} &\multicolumn{4}{c}{Change Detection (\%)}               \\ [0.01mm]
     & & P & R & IoU  & F1 
     & P & R & IoU  & F1   \\ [0.01mm]    
    \midrule  
   U-Net~\cite{ronneberger2015u} & 2015 
    &85.22 &85.26 &77.12 &85.24  
    &- &- &- &-
    \\[0.01mm]
    DeepLabv3+~\cite{chen2018encoder} & 2017
    &86.44 &88.68 &78.50 &87.55
    &- &- &- &-
    \\[0.01mm]
    HRNet~\cite{wang2020deep} & 2020 
    &85.87 &89.59 &78.22 &87.69 
     &- &- &- &-
    \\[0.01mm]
    SegFormer~\cite{xie2021segformer} & 2021 
    &86.82 &87.78 &78.02 &87.30
     &- &- &- &-
    \\ [0.01mm]
    DSAT-Net~\cite{zhang2023dsat} & 2023 
    &88.17 &88.72 &78.38 &88.44
    &- &- &- &-
    \\[0.01mm]
    \midrule  
    BIT~\cite{chen2021remote}& 2021     
    &- &- &- &-
    &71.16 &65.66 &51.81 &68.30
    \\[0.01mm]
   ChangeFormer~\cite{bandara2022transformer} & 2022       
    &- &- &- &-
    &74.48 &66.22 &53.89 &70.11
    \\ [0.01mm]      
    ChangerEx~\cite{fang2023changer}& 2023     
    &- &- &- &-
    &72.55 &\textbf{68.34}  &54.22 &70.38
    \\[0.01mm]     
    LCD-Net~\cite{liu2025lcd}  & 2024        
    &- &- &- &-
    &74.74 &67.62 &55.04 &71.00
    \\ [0.01mm]     
    IDA-SiamNet~\cite{li2024ida}  & 2024     
    &- &- &- &-
    &76.31 &66.75 &55.29 &71.21
    \\ [0.01mm]    
    \midrule   
    MSSFC-Net (Ours) &-
   & \textbf{88.89}  & \textbf{89.64} &  \textbf{79.10} &\textbf{89.26} & \textbf{76.92} &  67.28 & \textbf{56.38} & \textbf{71.78}
    \\[0.01mm]    
    \bottomrule
    \end{tabular}
}
\end{table*}

The visualization results of the dual-task are shown in Fig .\ref{fig:8}. The visualization results demonstrate the dual functions of our method: building extraction and change detection. Specifically, as shown in Fig .\ref{fig:8} (a), the MSSFC-Net method performs better than other methods in terms of the accurate depiction and representation of building edges. The deviation between the predicted contour and the actual building boundary is relatively small, and it has obvious advantages in maintaining the morphological details of buildings. Meanwhile, it effectively suppresses the pseudo-change regions caused by shadow interference factors.
 As shown in Fig .\ref{fig:8} (b), there are significant differences in the detection effects of various methods under different building scales. Although most models can identify the general building change regions, MSSFC-Net is particularly outstanding in detecting changes in densely built areas and can more accurately capture the change regions of dense buildings.
As shown in in Fig .\ref{fig:8} (c), it demonstrates the advanced ability to accurately identify buildings in low-contrast scenes. It can be seen from the figure that the recognition error of MSSFC-Net is significantly lower than that of other networks, while the false detection area of other networks is relatively large. The experimental results show that MSSFC-Net can describe the shape and location information of buildings relatively clearly, while other methods appear relatively blurry when describing the shape and edges of buildings after the changes. This result fully verifies the advanced ability of MSSFC-Net to accurately identify buildings in low-contrast scenes.
In addition, MSSFC-Net performs particularly well in scenarios with low false positive and false negative rates, while the recognition errors of other networks are relatively large.

\subsection{Ablation Studies and Analysis} 
In this section, we conduct a series of ablation experiments to provide a comprehensive analysis of the proposed MSSFC-Net method, aiming to investigate the underlying logic behind its superior performance.

\subsubsection{Effects of Model Component}

\begin{table}[t]
\centering
\caption{The impact of different model component designs (SSFC, MDFM, DMFE) was evaluated. The performance was validated on the WHU and LEVIR-CD test sets.}
\label{tab:whu-levircd}
\normalsize
  \begin{minipage}{0.5\textwidth} 
  \resizebox{\textwidth}{!}{
    \begin{tabular}{c c c|cc|cc}
    \toprule
    \multirow{2}{*}{MDFM}& \multirow{2}{*}{SSFC} & \multirow{2}{*}{DMFE} & \multicolumn{2}{c|}{WHU(\%)} &\multicolumn{2}{c}{LEVIR-CD(\%)}                                           
     \\ 
     & & & IoU  & F1 
     &  IoU  & F1   \\ 
    
    \midrule    
    $\times$ &$\times$ &$\times$  &89.62 &91.98 &83.58 &90.34
    \\    
    $\checkmark$ &$\times$ &$\times$ &90.67 &93.61 &84.39 &91.31
    \\ 
     $\checkmark$ &$\checkmark$ &$\times$   &91.08 &94.54 &85.02 &92.04 
    \\  
     $\checkmark$ &$\checkmark$ &$\checkmark$ &\textbf{92.27} &\textbf{95.82} &\textbf{85.78} &\textbf{92.48}
    \\ 
    \bottomrule
    \end{tabular}
}
\end{minipage}
\end{table}

\textbf{Effectiveness of the MDFM}: In MSSFC-Net, the MDFM module plays a crucial role in fusing bi-temporal features and embedding multi-scale semantic information into the difference features. Different from traditional feature concatenation or simple addition operations, this module adopts a strategy combining the calculation of the absolute value of feature differences and multi-level convolutional fusion.
The ablation experiment data (Tabel \ref{tab:whu-levircd}) shows that removing the MDFM will lead to a 1.05\% decrease in the IoU index of the WHU dataset and a 1.63\% decrease in the F1 value. The performance degradation is even more obvious on the LEVIR-CD dataset.
Through the parallel processing of multi-scale convolutional kernels (3×3, 5×5, 7×7), this module effectively separates the temporal change features from the static background information and suppresses the noise interference. On the WHU dataset, the IoU of MDFM reaches 90.67\% and the F1 value reaches 93.61\%. Similarly, MDFM also demonstrates excellent performance in the LEVIR-CD dataset.
It has successfully achieved the efficient fusion of the semantic information of bi-temporal features and the difference features, thereby significantly reducing the errors and improving the accuracy and robustness of change detection.

\textbf{Effectiveness of the SSFC}: 
The SSFC strategy generates more abundant and discriminative feature maps by modeling the complex relationships between spatial and spectral features. Tabel \ref{tab:whu-levircd} presents the ablation experiment results. The results indicate that the SSFC strategy we introduced, by efficiently modeling the dependencies between spatial and spectral features, automatically focuses on the target areas with significant changes. It effectively suppresses the redundant information in the auxiliary feature maps while retaining and enhancing the feature expressions useful for the task. The ablation experiment results in the table fully verify the effectiveness of the SSFC strategy in the multi-scale feature learning mechanism.
Experiments show that after adopting the SSFC strategy, the performance of the model in the change detection task is significantly improved, especially the ability to capture subtle changes in complex scenarios is evidently enhanced. On the WHU dataset, the IoU after adding SSFC reaches 90.67\% and the F1 value reaches 94.54\%. In the LEVIR-CD dataset, the IoU after adding SSFC reaches 85.02\% and the F1 value reaches 92.04\%. This indicates that the SSFC strategy can effectively distinguish buildings from background noise and reduce false positives and false negatives.
In addition, the SSFC strategy also demonstrates the ability to fuse multi-scale features, enabling it to perform excellently in target detection at different scales.

\textbf{Effectiveness of the DMFE}: The DMFE is designed to explore the interaction between bi-temporal features and difference features, enhance the change regions, and refine the difference features. Table \ref{tab:whu-levircd} lists the ablation experiment results of DMFE. When DMFE is removed, most of the indicators decrease. After adding DMFE, the performance improves.
On the WHU dataset, the IoU after adding DMFE reaches 92.27\% and the F1 value reaches 95.82\%. In the LEVIR-CD dataset, the IoU after adding DMFE reaches 85.78\% and the F1 value reaches 92.48\%. This reflects that our DMFE can promote the learning of bi-temporal features and difference features.
The presence of DMFE enables the model to delve deeper into the change regions, reducing false positives and false negatives. In addition, DMFE can effectively refine the features, enhance the change regions, and achieve good performance.

\subsubsection{Effects of Federated Training}

To evaluate the impact of each data segment on the model performance, we verified the performance enhancement mechanism of heterogeneous data modalities on the multi-task model through systematic pre-training experiments. In the experimental design, three dataset configuration strategies were adopted, which were divided into single-task training and dual-task training modes. Specifically, they included the Building Extraction dataset (BX), the Change Detection dataset (CD), and the Basic dataset (BX and CD).
After the pre-training stage was completed, a transfer learning strategy was employed to fine-tune the model for downstream task adaptation. As shown in Table \ref{tab:bx-cd}, the experimental results on the WHU dataset indicate that: for the training improvement effect of the unimodal task dataset, the training on the BX dataset increases the IoU of the building extraction task by 0.41\%, and the training on the CD dataset brings a 0.71\% gain in the F1-score of change detection. The jointly trained model achieves the optimal performance in both downstream tasks, with the IoU of building extraction reaching 92.27\% and the F1-score of change detection reaching 92.48\%.
In terms of task complementarity, we found that the BX modality strengthens the ability to extract spatial features, and the CD modality enhances the sensitivity to temporal differences. The two modalities achieve feature space sharing through the proposed cross-task knowledge distillation mechanism. The data confirms the potential semantic correlation between the building extraction and change detection tasks, and realizes the joint optimization of task-specific feature decoupling and shared parameters through the dynamic weight allocation module.

\begin{table}[t]
\centering
\caption{The performance was validated on the WHU and LEVIR-CD test sets. BX represents building extraction data, while CD refers to change detection data.}
\label{tab:bx-cd}
  \tiny
  \begin{minipage}{0.49\textwidth} 
  \renewcommand{\arraystretch}{0.9}
  \resizebox{\textwidth}{!}{
    \begin{tabular}{c|cc|cc}
    \toprule
    \multirow{2}{*}{Data} &\multicolumn{2}{c|}{WHU(\%)} &\multicolumn{2}{c}{LEVIR-CD(\%)}                                        
     \\ [0.01mm] 
     &   IoU  & F1 
     &  IoU  & F1   \\ [0.01mm] 
    \midrule    
    - &91.45 &94.97 &84.32 &91.87
    \\  [0.01mm]   
    BX &91.86 &95.32 &84.96 &92.05
    \\ [0.01mm] 
    CD  &92.03 &95.68 &85.42 &92.29 
    \\  [0.01mm] 
    BX \& CD &\textbf{92.27} &\textbf{95.82} &\textbf{85.78} &\textbf{92.48}
    \\ [0.01mm] 
    \bottomrule
    \end{tabular}
}
\end{minipage}
\end{table}

\subsubsection{Ablation study of the self-attention mechanism pooling layer in the SSFC module}

\begin{table}[t]
\centering
\caption{The ablation study of different pooling layers in the self-attention mechanism of the SSFC module on the LEVIR-CD dataset shows that the model achieves optimal performance when specific pooling strategy combinations are used. The corresponding results are highlighted in bold in the text.}
\label{tab:pool}
\tiny
\begin{minipage}{0.49\textwidth} 
\renewcommand{\arraystretch}{0.9} 
\resizebox{\textwidth}{!}{
    \begin{tabular}{c c|cc}
    \toprule
    \multirow{2}{*}{Query Features}& \multirow{2}{*}{Key Features} &\multicolumn{2}{c}{LEVIR-CD(\%)}                                           
     \\ [0.01mm] 
     & &IoU  & F1   \\ [0.01mm] 
    
    \midrule    
    - &-  &84.85 &91.73
    \\  [0.01mm]   
    Avg Pool &Avg Pool   &85.26 &91.91
     \\ [0.01mm]    
    Max Pool  &Max Pool   &85.04 &92.07
    \\ [0.01mm] 
    Max Pool  & Avg Pool  &85.32 &92.20 
    \\  [0.01mm] 
    \textbf{Avg Pool}  & \textbf{Max Pool}  &\textbf{85.78} &\textbf{92.48}
    \\ [0.01mm] 
    \bottomrule
    \end{tabular}
}
\end{minipage}
\end{table}

A systematic ablation experiment was carried out on the feature aggregation strategy in the DMFE module, with a focus on exploring the impact mechanism of pooling operations on multi-scale feature representation and computational efficiency. Combined experiments were conducted on the average pooling and max pooling operations for the query and key features.
In the Query Path, average pooling was applied to the query feature map. Through a smoothing operation, this helps to suppress high-frequency noise and enhance the robustness against minor deformations. In the Key Path, max pooling was used for the key feature map, which can strengthen the response intensity of local significant features.
As shown in Table \ref{tab:pool}, the experimental results indicated that this combined strategy led to a 0.93\% increase in IoU. Although single pooling strategies, namely pure average pooling and pure max pooling, could improve the IoU to some extent, they caused a decline in the positioning accuracy of edge details.
The experimental results demonstrated that the hybrid pooling was the optimal solution. The combined strategy of average pooling in the query path and max pooling in the key path achieved the best performance, with an IoU of 85.78\% and an F1 - score of 92.48\%. By complementing frequency-domain features, it reduced the positioning error of building boundaries and effectively suppressed noise.
The experimental results verified that the DMFE module, through a differentiated pooling strategy, realizes a collaborative optimization mechanism of global context awareness and local significant feature extraction. This provides a new technical path for efficient semantic segmentation in complex scenarios.

\subsubsection{The necessity of the SSFC strategy}

\begin{figure}[t]
    \centering
    \includegraphics[width=1\columnwidth]{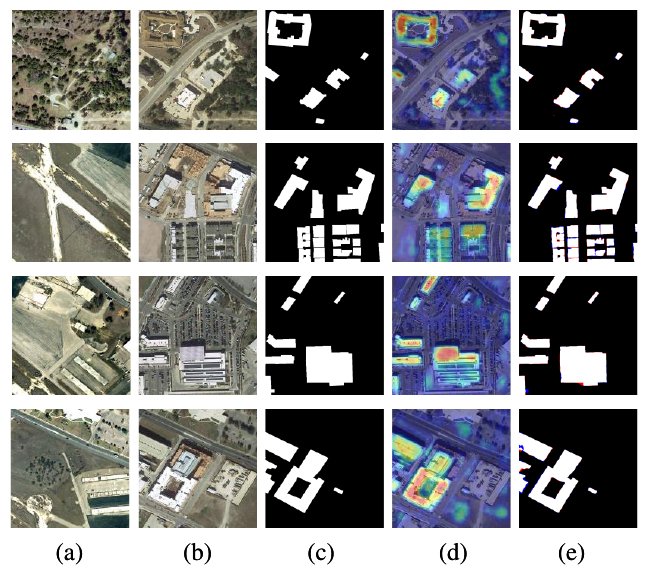}
    \caption{Feature activation after SSFC on the LEVIR-CD test set. (a) Pretem-poral image. (b) Posttemporal image. (c) Ground truth. (d) Feature activation map.(e) MSSFC-Net. Red and yellow in (c) denote higher attention values.}
    \label{fig:9}
\end{figure}

The SSFC strategy based on the collaboration of spatial-spectral features adopts a feature optimization mechanism. By establishing a spatial-spectral correlation model across dimensions, it significantly reduces redundant information while maintaining the feature expressiveness.
Specifically, the SSFC strategy employs an adaptive feature selection mechanism. Through establishing a multi-dimensional feature correlation matrix, it automatically identifies and strengthens the discriminative feature channels, and weakens the redundant responses, thus achieving feature optimization without adding new learnable parameters.
A spatial-spectral joint attention model is constructed to enable the network to autonomously focus on the regions with significant changes. As shown in Fig .\ref{fig:9}, in the feature activation visualization of the LEVIR-CD test set, the network attention guided by SSFC accurately covers the key areas such as the changes in building structures.
It is embedded into the DMFE module to form functional complementarity. On the one hand, it captures local details through multi-scale modeling, and on the other hand, it realizes global feature optimization.
By establishing a feature selection mechanism with a clear physical meaning, the SSFC strategy not only avoids the problem of dimensional explosion caused by simple fusion but also enhances the model's perception ability of change features. As shown in Fig .\ref{fig:9}, the visual analysis of the feature heatmap further verifies the advantages of the SSFC strategy in terms of interpretability, and its attention distribution is highly consistent with the change regions interpreted manually.

\section{Conclusion}\label{sec:con} 

In this work, we propose a spatial-spectral feature collaboration-based dual-task remote sensing image building extraction and change detection network, integrating both tasks into a unified framework. MSSFC-Net effectively integrates shallow-layer texture details with deep-layer semantic localization information, leveraging spectral and spatial features for information extraction. Additionally, by exploiting the relationship between difference features and bi-temporal features, the network significantly improves building extraction accuracy and subtle change detection capabilities. Extensive experiments validate the effectiveness of the proposed method. The proposed SS-BCNet demonstrates robust performance on three challenging benchmarks and achieves state-of-the-art results.This research will explore the feasibility of developing a lightweight version maintaining comparable performance in future studies.

{\small
\bibliographystyle{IEEEtran}
\bibliography{ref}

\begin{thebibliography}{10}
\providecommand{\url}[1]{#1}
\csname url@samestyle\endcsname
\providecommand{\newblock}{\relax}
\providecommand{\bibinfo}[2]{#2}
\providecommand{\BIBentrySTDinterwordspacing}{\spaceskip=0pt\relax}
\providecommand{\BIBentryALTinterwordstretchfactor}{4}
\providecommand{\BIBentryALTinterwordspacing}{\spaceskip=\fontdimen2\font plus
\BIBentryALTinterwordstretchfactor\fontdimen3\font minus \fontdimen4\font\relax}
\providecommand{\BIBforeignlanguage}[2]{{%
\expandafter\ifx\csname l@#1\endcsname\relax
\typeout{** WARNING: IEEEtran.bst: No hyphenation pattern has been}%
\typeout{** loaded for the language `#1'. Using the pattern for}%
\typeout{** the default language instead.}%
\else
\language=\csname l@#1\endcsname
\fi
#2}}
\providecommand{\BIBdecl}{\relax}
\BIBdecl

\bibitem{dai2020object}
C.~Dai, Z.~Zhang, and D.~Lin, ``An object-based bidirectional method for integrated building extraction and change detection between multimodal point clouds,'' \emph{Remote sensing}, vol.~12, no.~10, p. 1680, 2020.

\bibitem{mohammadian2023siamixformer}
A.~Mohammadian and F.~Ghaderi, ``Siamixformer: A fully-transformer siamese network with temporal fusion for accurate building detection and change detection in bi-temporal remote sensing images,'' \emph{International Journal of Remote Sensing}, vol.~44, no.~12, pp. 3660--3678, 2023.

\bibitem{lu2024bi}
W.~Lu, L.~Wei, and M.~Nguyen, ``Bi-temporal attention transformer for building change detection and building damage assessment,'' \emph{IEEE Journal of Selected Topics in Applied Earth Observations and Remote Sensing}, 2024.

\bibitem{gong2023context}
M.~Gong, T.~Liu, M.~Zhang, Q.~Zhang, D.~Lu, H.~Zheng, and F.~Jiang, ``Context--content collaborative network for building extraction from high-resolution imagery,'' \emph{Knowledge-Based Systems}, vol. 263, p. 110283, 2023.

\bibitem{lv2023spatial}
Z.~Lv, M.~Zhang, W.~Sun, J.~A. Benediktsson, T.~Lei, and N.~Falco, ``Spatial-contextual information utilization framework for land cover change detection with hyperspectral remote sensed images,'' \emph{IEEE Transactions on Geoscience and Remote Sensing}, vol.~61, pp. 1--11, 2023.

\bibitem{wang2024rsbuilding}
M.~Wang, K.~Chen, L.~Su, C.~Yan, S.~Xu, H.~Zhang, P.~Yuan, X.~Jiang, and B.~Zhang, ``Rsbuilding: Towards general remote sensing image building extraction and change detection with foundation model,'' \emph{arXiv preprint arXiv:2403.07564}, 2024.

\bibitem{wang2024sdsnet}
X.~Wang, M.~Tian, Z.~Zhang, K.~He, S.~Wang, Y.~Liu, and Y.~Dong, ``Sdsnet: Building extraction in high-resolution remote sensing images using a deep convolutional network with cross-layer feature information interaction filtering,'' \emph{Remote Sensing}, vol.~16, no.~1, p. 169, 2024.

\bibitem{shen2023pbsl}
F.~Shen, X.~Shu, X.~Du, and J.~Tang, ``Pedestrian-specific bipartite-aware similarity learning for text-based person retrieval,'' in \emph{Proceedings of the 31th ACM International Conference on Multimedia}, 2023.

\bibitem{sun2024g2ldie}
J.~Sun, W.~He, and H.~Zhang, ``G2ldie: Global-to-local dynamic information enhancement framework for weakly supervised building extraction from remote sensing images,'' \emph{IEEE Transactions on Geoscience and Remote Sensing}, 2024.

\bibitem{wang2022building}
L.~Wang, S.~Fang, X.~Meng, and R.~Li, ``Building extraction with vision transformer,'' \emph{IEEE Transactions on Geoscience and Remote Sensing}, vol.~60, pp. 1--11, 2022.

\bibitem{huang2023easy}
H.~Huang, J.~Liu, and R.~Wang, ``Easy-net: A lightweight building extraction network based on building features,'' \emph{IEEE Transactions on Geoscience and Remote Sensing}, 2023.

\bibitem{li2024ida}
Y.-C. Li, S.~Lei, N.~Liu, H.-C. Li, and Q.~Du, ``Ida-siamnet: Interactive-and dynamic-aware siamese network for building change detection,'' \emph{IEEE Transactions on Geoscience and Remote Sensing}, 2024.

\bibitem{pang2023sfgt}
S.~Pang, J.~Lan, Z.~Zuo, and J.~Chen, ``Sfgt-cd: Semantic feature-guided building change detection from bitemporal remote sensing images with transformers,'' \emph{IEEE Geoscience and Remote Sensing Letters}, 2023.

\bibitem{shen2024boosting}
F.~Shen, H.~Ye, S.~Liu, J.~Zhang, C.~Wang, X.~Han, and W.~Yang, ``Boosting consistency in story visualization with rich-contextual conditional diffusion models,'' \emph{arXiv preprint arXiv:2407.02482}, 2024.

\bibitem{yang2024sedanet}
Y.~Yang, T.~Chen, T.~Lei, B.~Du, A.~K. Nandi, and A.~Plaza, ``Sedanet: A new siamese ensemble difference attention network for building change detection in remotely sensed images,'' \emph{IEEE Transactions on Geoscience and Remote Sensing}, 2024.

\bibitem{bandara2022transformer}
W.~G.~C. Bandara and V.~M. Patel, ``A transformer-based siamese network for change detection,'' in \emph{IGARSS 2022-2022 IEEE International Geoscience and Remote Sensing Symposium}.\hskip 1em plus 0.5em minus 0.4em\relax IEEE, 2022, pp. 207--210.

\bibitem{chen2021remote}
H.~Chen, Z.~Qi, and Z.~Shi, ``Remote sensing image change detection with transformers,'' \emph{IEEE Transactions on Geoscience and Remote Sensing}, vol.~60, pp. 1--14, 2021.

\bibitem{shangguan2023attention}
Y.~Shangguan, J.~Li, Y.~Liu, F.~Zhang, and C.~Zhang, ``Attention filtering network based on branch transformer for change detection in remote sensing images,'' \emph{IEEE Transactions on Geoscience and Remote Sensing}, 2023.

\bibitem{lv2023hierarchical}
Z.~Lv, J.~Liu, W.~Sun, T.~Lei, J.~A. Benediktsson, and X.~Jia, ``Hierarchical attention feature fusion-based network for land cover change detection with homogeneous and heterogeneous remote sensing images,'' \emph{IEEE Transactions on Geoscience and Remote Sensing}, vol.~61, pp. 1--15, 2023.

\bibitem{zhang2023self}
M.~Zhang, H.~Zheng, M.~Gong, Y.~Wu, H.~Li, and X.~Jiang, ``Self-structured pyramid network with parallel spatial-channel attention for change detection in vhr remote sensed imagery,'' \emph{Pattern Recognition}, vol. 138, p. 109354, 2023.

\bibitem{noman2024changebind}
M.~Noman, M.~Fiaz, and H.~Cholakkal, ``Changebind: A hybrid change encoder for remote sensing change detection,'' in \emph{IGARSS 2024-2024 IEEE International Geoscience and Remote Sensing Symposium}.\hskip 1em plus 0.5em minus 0.4em\relax IEEE, 2024, pp. 8417--8422.

\bibitem{wang2023align}
S.~Wang, Y.~Li, M.~Xie, M.~Chi, Y.~Wang, C.~Wang, and W.~Zhu, ``Align, perturb and decouple: Toward better leverage of difference information for rsi change detection,'' \emph{arXiv preprint arXiv:2305.18714}, 2023.

\bibitem{hu2023robust}
J.~Hu, Z.~Huang, F.~Shen, D.~He, and Q.~Xian, ``A rubust method for roof extraction and height estimation,'' in \emph{IGARSS 2023-2023 IEEE International Geoscience and Remote Sensing Symposium}.\hskip 1em plus 0.5em minus 0.4em\relax IEEE, 2023.

\bibitem{huang2023deep}
L.~Huang, B.~Jiang, S.~Lv, Y.~Liu, and Y.~Fu, ``Deep learning-based semantic segmentation of remote sensing images: A survey,'' \emph{IEEE Journal of Selected Topics in Applied Earth Observations and Remote Sensing}, 2023.

\bibitem{yan2024enhancing}
K.~Yan, F.~Shen, and Z.~Li, ``Enhancing landslide segmentation with guide attention mechanism and fast fourier transformer,'' in \emph{International Conference on Intelligent Computing}.\hskip 1em plus 0.5em minus 0.4em\relax Springer, 2024, pp. 296--307.

\bibitem{ronneberger2015u}
O.~Ronneberger, P.~Fischer, and T.~Brox, ``U-net: Convolutional networks for biomedical image segmentation,'' in \emph{Medical image computing and computer-assisted intervention--MICCAI 2015: 18th international conference, Munich, Germany, October 5-9, 2015, proceedings, part III 18}.\hskip 1em plus 0.5em minus 0.4em\relax Springer, 2015, pp. 234--241.

\bibitem{zhou2018unet++}
Z.~Zhou, M.~M. Rahman~Siddiquee, N.~Tajbakhsh, and J.~Liang, ``Unet++: A nested u-net architecture for medical image segmentation,'' in \emph{Deep Learning in Medical Image Analysis and Multimodal Learning for Clinical Decision Support: 4th International Workshop, DLMIA 2018, and 8th International Workshop, ML-CDS 2018, Held in Conjunction with MICCAI 2018, Granada, Spain, September 20, 2018, Proceedings 4}.\hskip 1em plus 0.5em minus 0.4em\relax Springer, 2018, pp. 3--11.

\bibitem{oktay2018attention}
O.~Oktay, J.~Schlemper, L.~L. Folgoc, M.~Lee, M.~Heinrich, K.~Misawa, K.~Mori, S.~McDonagh, N.~Y. Hammerla, B.~Kainz \emph{et~al.}, ``Attention u-net: Learning where to look for the pancreas,'' \emph{arXiv preprint arXiv:1804.03999}, 2018.

\bibitem{dosovitskiy2020image}
A.~Dosovitskiy, ``An image is worth 16x16 words: Transformers for image recognition at scale,'' \emph{arXiv preprint arXiv:2010.11929}, 2020.

\bibitem{wang2022uformer}
Z.~Wang, X.~Cun, J.~Bao, W.~Zhou, J.~Liu, and H.~Li, ``Uformer: A general u-shaped transformer for image restoration,'' in \emph{Proceedings of the IEEE/CVF conference on computer vision and pattern recognition}, 2022, pp. 17\,683--17\,693.

\bibitem{daudt2018fully}
R.~C. Daudt, B.~Le~Saux, and A.~Boulch, ``Fully convolutional siamese networks for change detection,'' in \emph{2018 25th IEEE international conference on image processing (ICIP)}.\hskip 1em plus 0.5em minus 0.4em\relax IEEE, 2018, pp. 4063--4067.

\bibitem{fang2021snunet}
S.~Fang, K.~Li, J.~Shao, and Z.~Li, ``Snunet-cd: A densely connected siamese network for change detection of vhr images,'' \emph{IEEE Geoscience and Remote Sensing Letters}, vol.~19, pp. 1--5, 2021.

\bibitem{ying2024dgma}
Z.~Ying, Z.~Tan, Y.~Zhai, X.~Jia, W.~Li, J.~Zeng, A.~Genovese, V.~Piuri, and F.~Scotti, ``Dgma 2-net: A difference-guided multiscale aggregation attention network for remote sensing change detection,'' \emph{IEEE Transactions on Geoscience and Remote Sensing}, 2024.

\bibitem{li2022transunetcd}
Q.~Li, R.~Zhong, X.~Du, and Y.~Du, ``Transunetcd: A hybrid transformer network for change detection in optical remote-sensing images,'' \emph{IEEE Transactions on Geoscience and Remote Sensing}, vol.~60, pp. 1--19, 2022.

\bibitem{gedara2022transformer}
W.~Gedara, C.~Bandara, and V.~M. Patel, ``A transformer-based siamese network for change detection,'' \emph{pp. arXiv-2201}, 2022.

\bibitem{shen2023advancing}
F.~Shen, H.~Ye, J.~Zhang, C.~Wang, X.~Han, and W.~Yang, ``Advancing pose-guided image synthesis with progressive conditional diffusion models,'' \emph{arXiv preprint arXiv:2310.06313}, 2023.

\bibitem{shen2024imagdressing}
F.~Shen, X.~Jiang, X.~He, H.~Ye, C.~Wang, X.~Du, Z.~Li, and J.~Tang, ``Imagdressing-v1: Customizable virtual dressing,'' \emph{arXiv preprint arXiv:2407.12705}, 2024.

\bibitem{shen2024imagpose}
F.~Shen and J.~Tang, ``Imagpose: A unified conditional framework for pose-guided person generation,'' in \emph{The Thirty-eighth Annual Conference on Neural Information Processing Systems}, 2024.

\bibitem{vaswani2017attention}
A.~Vaswani, ``Attention is all you need,'' \emph{Advances in Neural Information Processing Systems}, 2017.

\bibitem{hu2018squeeze}
J.~Hu, L.~Shen, and G.~Sun, ``Squeeze-and-excitation networks,'' in \emph{Proceedings of the IEEE conference on computer vision and pattern recognition}, 2018, pp. 7132--7141.

\bibitem{woo2018cbam}
S.~Woo, J.~Park, J.-Y. Lee, and I.~S. Kweon, ``Cbam: Convolutional block attention module,'' in \emph{Proceedings of the European conference on computer vision (ECCV)}, 2018, pp. 3--19.

\bibitem{lei2021difference}
T.~Lei, J.~Wang, H.~Ning, X.~Wang, D.~Xue, Q.~Wang, and A.~K. Nandi, ``Difference enhancement and spatial--spectral nonlocal network for change detection in vhr remote sensing images,'' \emph{IEEE Transactions on Geoscience and Remote Sensing}, vol.~60, pp. 1--13, 2021.

\bibitem{li2023cbanet}
Y.~Li, J.~Ren, Y.~Yan, Q.~Liu, P.~Ma, A.~Petrovski, and H.~Sun, ``Cbanet: An end-to-end cross-band 2-d attention network for hyperspectral change detection in remote sensing,'' \emph{IEEE Transactions on Geoscience and Remote Sensing}, vol.~61, pp. 1--11, 2023.

\bibitem{guo2022beyond}
M.-H. Guo, Z.-N. Liu, T.-J. Mu, and S.-M. Hu, ``Beyond self-attention: External attention using two linear layers for visual tasks,'' \emph{IEEE Transactions on Pattern Analysis and Machine Intelligence}, vol.~45, no.~5, pp. 5436--5447, 2022.

\bibitem{szegedy2015going}
C.~Szegedy, W.~Liu, Y.~Jia, P.~Sermanet, S.~Reed, D.~Anguelov, D.~Erhan, V.~Vanhoucke, and A.~Rabinovich, ``Going deeper with convolutions,'' in \emph{Proceedings of the IEEE conference on computer vision and pattern recognition}, 2015, pp. 1--9.

\bibitem{lei2023ultralightweight}
T.~Lei, X.~Geng, H.~Ning, Z.~Lv, M.~Gong, Y.~Jin, and A.~K. Nandi, ``Ultralightweight spatial--spectral feature cooperation network for change detection in remote sensing images,'' \emph{IEEE Transactions on Geoscience and Remote Sensing}, vol.~61, pp. 1--14, 2023.

\bibitem{fu2019dual}
J.~Fu, J.~Liu, H.~Tian, Y.~Li, Y.~Bao, Z.~Fang, and H.~Lu, ``Dual attention network for scene segmentation,'' in \emph{Proceedings of the IEEE/CVF conference on computer vision and pattern recognition}, 2019, pp. 3146--3154.

\bibitem{ji2018fully}
S.~Ji, S.~Wei, and M.~Lu, ``Fully convolutional networks for multisource building extraction from an open aerial and satellite imagery data set,'' \emph{IEEE Transactions on geoscience and remote sensing}, vol.~57, no.~1, pp. 574--586, 2018.

\bibitem{zhang2020deeply}
C.~Zhang, P.~Yue, D.~Tapete, L.~Jiang, B.~Shangguan, L.~Huang, and G.~Liu, ``A deeply supervised image fusion network for change detection in high resolution bi-temporal remote sensing images,'' \emph{ISPRS Journal of Photogrammetry and Remote Sensing}, vol. 166, pp. 183--200, 2020.

\bibitem{pang2023detecting}
C.~Pang, J.~Wu, J.~Ding, C.~Song, and G.-S. Xia, ``Detecting building changes with off-nadir aerial images,'' \emph{Science China Information Sciences}, vol.~66, no.~4, p. 140306, 2023.

\bibitem{chen2018encoder}
L.-C. Chen, Y.~Zhu, G.~Papandreou, F.~Schroff, and H.~Adam, ``Encoder-decoder with atrous separable convolution for semantic image segmentation,'' in \emph{Proceedings of the European conference on computer vision (ECCV)}, 2018, pp. 801--818.

\bibitem{wang2020deep}
J.~Wang, K.~Sun, T.~Cheng, B.~Jiang, C.~Deng, Y.~Zhao, D.~Liu, Y.~Mu, M.~Tan, X.~Wang \emph{et~al.}, ``Deep high-resolution representation learning for visual recognition,'' \emph{IEEE transactions on pattern analysis and machine intelligence}, vol.~43, no.~10, pp. 3349--3364, 2020.

\bibitem{xie2021segformer}
E.~Xie, W.~Wang, Z.~Yu, A.~Anandkumar, J.~M. Alvarez, and P.~Luo, ``Segformer: Simple and efficient design for semantic segmentation with transformers,'' \emph{Advances in neural information processing systems}, vol.~34, pp. 12\,077--12\,090, 2021.

\bibitem{zhu2020map}
Q.~Zhu, C.~Liao, H.~Hu, X.~Mei, and H.~Li, ``Map-net: Multiple attending path neural network for building footprint extraction from remote sensed imagery,'' \emph{IEEE Transactions on Geoscience and Remote Sensing}, vol.~59, no.~7, pp. 6169--6181, 2020.

\bibitem{zhou2022bomsc}
Y.~Zhou, Z.~Chen, B.~Wang, S.~Li, H.~Liu, D.~Xu, and C.~Ma, ``Bomsc-net: Boundary optimization and multi-scale context awareness based building extraction from high-resolution remote sensing imagery,'' \emph{IEEE Transactions on Geoscience and Remote Sensing}, vol.~60, pp. 1--17, 2022.

\bibitem{xu2023bctnet}
L.~Xu, Y.~Li, J.~Xu, Y.~Zhang, and L.~Guo, ``Bctnet: Bi-branch cross-fusion transformer for building footprint extraction,'' \emph{IEEE Transactions on Geoscience and Remote Sensing}, vol.~61, pp. 1--14, 2023.

\bibitem{zhang2023dsat}
R.~Zhang, Z.~Wan, Q.~Zhang, and G.~Zhang, ``Dsat-net: Dual spatial attention transformer for building extraction from aerial images,'' \emph{IEEE Geoscience and Remote Sensing Letters}, 2023.

\bibitem{wang2023sdsnet}
X.~Wang, M.~Tian, Z.~Zhang, K.~He, S.~Wang, Y.~Liu, and Y.~Dong, ``Sdsnet: Building extraction in high-resolution remote sensing images using a deep convolutional network with cross-layer feature information interaction filtering,'' \emph{Remote Sensing}, vol.~16, no.~1, p. 169, 2023.

\bibitem{chen2024sau}
M.~Chen, T.~Mao, J.~Wu, R.~Du, B.~Zhao, and L.~Zhou, ``Sau-net: A novel network for building extraction from high-resolution remote sensing images by reconstructing fine-grained semantic features,'' \emph{IEEE Journal of Selected Topics in Applied Earth Observations and Remote Sensing}, vol.~17, pp. 6747--6761, 2024.

\bibitem{yang2024csa}
D.~Yang, X.~Gao, Y.~Yang, M.~Jiang, K.~Guo, B.~Liu, S.~Li, and S.~Yu, ``Csa-net: Complex scenarios adaptive network for building extraction for remote sensing images,'' \emph{IEEE Journal of Selected Topics in Applied Earth Observations and Remote Sensing}, 2024.

\bibitem{fang2023changer}
S.~Fang, K.~Li, and Z.~Li, ``Changer: Feature interaction is what you need for change detection,'' \emph{IEEE Transactions on Geoscience and Remote Sensing}, vol.~61, pp. 1--11, 2023.

\bibitem{feng2024sgnet}
J.~Feng, X.~Yang, and Z.~Gu, ``Sgnet: A transformer-based semantic-guided network for building change detection,'' \emph{IEEE Journal of Selected Topics in Applied Earth Observations and Remote Sensing}, 2024.

\bibitem{liu2025lcd}
W.~Liu, J.~Li, H.~Wang, R.~Tan, Y.~Fu, and Q.~Tian, ``Lcd-net: A lightweight remote sensing change detection network combining feature fusion and gating mechanism,'' \emph{IEEE Journal of Selected Topics in Applied Earth Observations and Remote Sensing}, 2025.

\end{thebibliography}
}

\vfill
\end{document}